\documentclass[twoside]{article}

\usepackage[accepted]{aistats2024}
\newcommand{\std}[1]{\tiny{$\pm$#1}}

\usepackage{microtype}
\usepackage{graphicx}
\usepackage{subfigure}
\usepackage{booktabs} 
\usepackage{bm}

\usepackage{hyperref}
\usepackage{url}
\usepackage{multirow}
\usepackage[flushleft]{threeparttable}
\usepackage{longtable}
\usepackage{supertabular}
\usepackage{colortbl}
\usepackage{xspace}
\usepackage[linesnumbered,ruled,vlined]{algorithm2e}
\usepackage{wrapfig}
\usepackage[table]{xcolor}
\usepackage{xcolor}
\usepackage{cancel}
\usepackage{amsthm}
\usepackage{amsmath}

\usepackage{natbib}

\definecolor{maroon}{cmyk}{0,0.1,0.01,0.01}
\definecolor{blue}{cmyk}{0.95,0.0,0.2,0.2}
\definecolor{yellow}{cmyk}{0.01,0.0,0.2,0.01}
\definecolor{lightblue}{cmyk}{0.1,0.0,0.02,0.02}

\begin{document}

\runningtitle{Enhancing Uncertainty Quantification
and Interpretability with Graph Functional Neural Process}

\runningauthor{Lingkai Kong*, Haotian Sun*, Yuchen Zhuang, Haorui Wang, Wenhao Mu, Chao Zhang}

\twocolumn[

\aistatstitle{Two Birds with One Stone: Enhancing Uncertainty Quantification and  Interpretability with Graph  Functional Neural Process}

\aistatsauthor{ Lingkai Kong* \And Haotian Sun* \And  Yuchen Zhuang \And  Haorui Wang}
\vspace{0.5em}
\aistatsauthor{Wenhao Mu \And Chao Zhang}
\vspace{0.3em}
\aistatsaddress{ School of Computational Science and Engineering \\
Georgia Institute of Technology} ]

\newcommand{\ours}{\textsc{GraphFNP}\xspace}
\newcommand{\etal}{\textit{et al}. }
\newcommand{\ie}{\textit{i}.\textit{e}.,}
\newcommand{\eg}{\textit{e}.\textit{g}.}

\begin{abstract}
  Graph neural networks (GNNs) are powerful tools on graph data. However, their predictions are mis-calibrated and lack interpretability, limiting their adoption in critical applications.
  To address this issue, we propose a new uncertainty-aware and interpretable graph classification model that combines graph  functional neural process and graph generative model. The core of our method is to assume a set of latent rationales which can be mapped to a probabilistic embedding space; the predictive distribution of the classifier is conditioned on such rationale embeddings by learning a stochastic correlation matrix. The graph generator serves to decode the graph structure of the rationales from the embedding space for model interpretability.
  For efficient model training, we adopt an alternating optimization procedure which mimics the well known Expectation-Maximization (EM) algorithm. The proposed method is general and can be applied to any existing GNN architecture.
  Extensive experiments on five graph classification datasets demonstrate that our framework outperforms state-of-the-art methods in both uncertainty quantification and GNN interpretability.  We also conduct case studies to show that the decoded rationale structure can provide meaningful explanations.
\end{abstract}

\section{Introduction}
  Graph neural networks (GNNs) \citep{kipf2016semi, velivckovic2018graph, hamilton2017inductive, li2016gated} have been successful in various graph analytic tasks, such as graph classification, node classification and link prediction. GNNs provide a flexible framework to learn node representations with the message passing scheme, which aggregates vector representations from their topological neighborhoods. Compared with traditional graph mining techniques, GNNs transform the graph from the discrete graph space into the continuous embedding space that is easier to optimize for downstream tasks. Moreover, they can leverage the representation power of deep neural networks (DNNs) to learn complex input-output mapping functions and thus can achieve high accuracy across many tasks.

Many graph applications demand not only accurate but also \textit{uncertainty-aware} and \textit{explainable} predictions.
These two features are crucial for understanding and building trust in GNN predictions.
First, the absence of uncertainty estimates can result in unreliable probabilistic predictions and failure in practice~\citep{kong2023uncertainty}. 
For example, in molecular property prediction~\citep{wieder2020compact, feinberg2018potentialnet}, highly parameterized GNNs are vulnerable to overfitting to training scaffolds and may produce incorrect and poorly calibrated predictions for new scaffolds without uncertainty quantification.
Second,
GNNs are commonly used as black-box predictors, lacking explanations for their predictions.
For example, it is important to
understand which chemical groups in a molecular graph contribute to the predictions in molecular property prediction \citep{amara2023explaining}. However, the black-box nature of current GNNs makes it difficult to verify if their working mechanisms align with real-world chemical rules, reducing trust in their predictions and limiting their adoption in critical applications.

The challenge of providing uncertainty-aware and explainable predictions with GNNs remains unresolved.
Recent studies \citep{wang2021confident} have shown that GNNs are poor at quantifying predictive uncertainty and miscalibrated in their predictions.
One natural idea to remedy this issue is to apply existing uncertainty quantification techniques for GNNs, such as model ensembling and Bayesian neural networks (BNNs).
Model ensembling trains multiple deep neural networks (DNNs) with different initializations and ensembles their predictions for uncertainty quantification\citep{lakshminarayanan2017simple,ganaie2021ensemble}, but this approach incurs significant computation cost.
BNN \citep{welling2011bayesian, louizos2017multiplicative, ritter2018a, blundell2015weight,li2015preconditioned, zhang2019cyclical} quantify uncertainty by imposing probability distributions over model parameters, but the exact inference of the posterior distribution and proper specification of prior distributions for uninterpretable GNN parameters remain difficult.
With regard to interpretability, most existing graph interpretable methods \citep{ying2019gnnexplainer, luo2020parameterized} provide sample-level explanations, but these are too specific and difficult to generalize.
Instead, model-level interpretations, which aim to explain the overall behavior of the model by uncovering the key patterns or substructures driving predictions, are more general and require less human supervision.
However, generating model-level explanations for graphs remains an underexplored area.

We present a new uncertainty-aware and explainable graph classification approach that combines graph functional neural process and graph generative model. This framework has the following three capabilities: (1) quantifying predictive uncertainty directly from the functional space, (2) generating model-level rationales for interpretability, and (3) being applicable to any GNN architecture. The method assumes the existence of latent model-level rationales that can be mapped to distributions in the embedding space and predicts the class based on these stochastic rationale embeddings, providing a natural way to quantify uncertainty. The classifier, inspired by the functional neural process (FNP) \citep{louizos2019functional}, models the relationship between rationales and training graphs in the shared latent embedding space by learning a stochastic correlation matrix and generates the final predictive distribution based on correlated rationales. Additionally, the framework includes an autoregressive graph generator that produces the graph structure of the rationale embeddings for interpretability.

We  have conducted  thorough experiments to evaluate our proposed  model on five graph classification datasets. Our model consistently outperforms existing state-of-art (SOTA) methods in both uncertainty quantification and model interpretability. Specifically, \ours outperforms the strongest baseline by up to $4.8\%$ in terms of expected calibration error and meanwhile  maintains competitive predictive performance. To quantitatively evaluate the learned rationales, we apply a KNN-based classifier by computing the distance between the rationales and input graphs in the embedding space; \ours outperforms SOTA methods by up to $10.8\%$ in terms of F1 score. We also qualitatively visualize the decoded rationale structure to show that they do contain critical substructure for the class and align with the real-world rules.

\vspace{-0.7em}

\begin{figure*}
  \centering
  \includegraphics[width=0.95\textwidth]{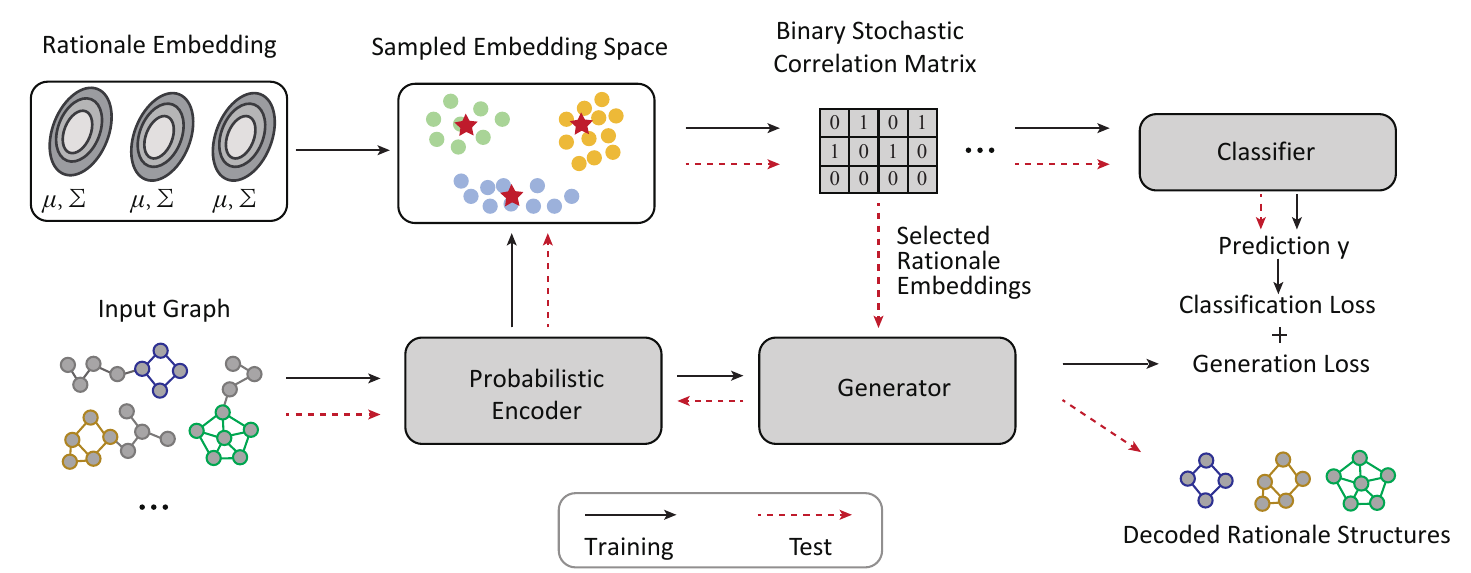}
  \caption{\label{fig:generative process} The overall generative process of the proposed
    framework. We assume a set of latent model-level rationales which can be mapped into a probabilistic embedding space; the graph classifier is conditioned on such rationale embeddings and graph embeddings through a stochastic correlation matrix. The graph generator is used to obtain the graph structure of the selected rationales from the latent embedding space.
  }
  \vspace{-1em}
\end{figure*}

\section{Notations and Problem Definition}
\vspace{-0.7em}
We focus on the graph classification problem in this paper. Let $D=\{(G^D_i,y^D_i)
\}_{i=1}^N$ be the set of labeled training data. $G=(\mathcal{V},\mathcal{E})$
represents a graph with $\mathcal{V}=\{v_1,v_2, \cdots, v_{n} \}$ denoting
the node set and $\mathcal{E}\in \mathcal{V} \times \mathcal{V}$ denoting the edge set.  The numbers of nodes and edges are denoted by $n$ and $m$, respectively. The nodes in $\mathcal{V}$ are associated with the $d$-dimensional features, denoted  as by $\mathbf{X}\in\mathcal{R}^{n\times d}$. $y=\{1, 2, \cdots, K\}$ denotes the category of the corresponding whole graph and $K$ the total number of categories. Taking the molecular property prediction as an example, $\mathcal{V}$ is the set of atoms; $\mathcal{E}$ is the set of bonds; $\mathbf{X}$ is the one-hot encoding of the atom type.

The graph classification problem involves learning a model $\mathcal{M}_{\theta}$, parameterized by $\theta$, that can categorize a graph $G$ into different classes. The model outputs a predicted category $\hat{y}$ and its corresponding confidence $\hat{p}$, such that $\mathcal{M}_{\theta}(G) \rightarrow (\hat{y},\hat{p})$. 
Traditional GNNs only provide point estimates and lack interpretability, making them unsuitable for  safety-critical applications. Our goal is to train a model that, given an unseen graph $G^{*}$ at test time, can provide both (1) a well-calibrated predictive distribution $p(y^{*}|G^{*})$ and (2) a model-level rationale $G^R$ that explains the crucial patterns that led to the prediction.

\noindent
\textbf{Calibrated Uncertainty Estimates}:
A well-calibrated predictive model should have confidence levels in its predictions that align with the actual accuracy of those predictions. For instance, if 100 data points are predicted with a confidence of 0.8, we expect 80 of them to be correctly classified. The calibration error of the predictive model, as defined by \citet{guo2017calibration, kong2020calibrated}, measures the discrepancy between the model's confidence in its predictions and the actual accuracy of those predictions. Given a confidence level $p \in [0, 1]$, the calibration error is given by:
\begin{equation}
  \mathcal{E}_{p}=|\mathcal{P}(\hat{y}=y|\hat{p}=p) - p|,
  \label{ce}
\end{equation}
where $\mathcal{P}(\hat{y}=y|\hat{p}=p)$ is the probability that the predicted class $\hat{y}$ is the same as the true class $y$, given the predicted confidence level $\hat{p}=p$. A model with perfect predictive uncertainty should satisfy $\mathcal{P}(\hat{y}=y|\hat{p}=p) = p$ for all $p \in [0, 1]$.

\noindent
\textbf{Model-Level Rationales for Interpretability}:
In graph classification, model-level rationales refer to a collection of subgraph structures, ${(G^R_{i}, y^R_{i})}_{i=1}^{|R|}$, where each subgraph $G_i^R$ represents a key pattern associated with its corresponding category $y_i^R$. For instance, in molecular property prediction, a molecule containing a $NO_2$ substructure often has mutagenic properties. By identifying such salient patterns, practitioners can verify if the model aligns with real-world rules and gain actionable insights.
Note that, unlike sample-level rationales \citep{ying2019gnnexplainer, luo2020parameterized,schlichtkrull2020interpreting}, which are defined as subgraphs of individual input graphs, model-level rationales are not specific to any particular graph.

\vspace{-0.8em}

\section{Methodology}
\vspace{-0.8em}
\subsection{Model Overview}
\vspace{-0.5em}
Our objective is to learn a predictive probability distribution $p(y|G)$ from the training data $D$ and provide model-level rationales to explain the predictions. To this end, we propose a non-parametric probabilistic generative process that jointly models the graph generation and classification procedures. This framework is capable of: (1) directly estimating predictive uncertainty in the functional space and (2) identifying the relationship between model predictions and generated rationales for better interpretability.

Our framework learns a set of  rationale embeddings from the continuous embedding space.
The rationale embeddings hold crucial information for the classification task, and so the predictive distribution $p(y|G)$ is based on them, which transforms the problem into learning the correlation between rationale and graph embeddings. Inspired by the functional neural process (FNP) \citep{louizos2019functional}, we represent these correlations with a binary stochastic matrix. Furthermore, we include a graph generative model to derive the graph structure of rationales from their embeddings.

Our method has three key components (Fig.~\ref{fig:generative process}):

(1) \textbf{Probabilistic Rationale Embedding}: To learn the rationale embeddings $\mathbf{Z}^R=\{ \mathbf{z}_i^R\}_{i=1}^{|R|}$ without having access to rationales and to make the learning process differentiable, we opt to learn a set of rationale embeddings from the latent embedding space. To capture the uncertainty in the embeddings, we represent them as high-dimensional Gaussian random variables. The input graphs are transformed into the same space through a GNN encoder.

(2) \textbf{Stochastic Correlation Matrix}: The correlation matrix $\mathbf{C}$ models the relationship between training graphs and rationales in the embedding space. It serves a similar role as a kernel function in Gaussian processes (GPs) for non-parametric uncertainty estimation. The correlated rationales reflect the crucial substructures emphasized by the model for prediction. The final predictive distribution is parameterized with two stochastic latent variables:
(a) The local rationale embedding $\mathbf{U}^D=\{\mathbf{u}^D_i\}_{i=1}^N$, which summarize correlated rationale embeddings. (b) The graph embedding $\mathbf{Z}^D=\{\mathbf{z}^D_i\}_{i=1}^N$, which captures embedding uncertainty and provides novel information not present in the rationales.

(3) \textbf{Graph Generative Model for Model Interpretability}: This component is crucial for model interpretability as it allows us to obtain the rationale structure from the learned rationale embeddings using the decoder after training. We design the graph generator as a simple and flexible autoregressive model. To incorporate information from the latent embedding space, our generation procedure is dependent on the graph/rationale embeddings.

Combing all components, we arrive at the following generative process for the training data:
\begin{align}
    &p(\mathbf{y}^D|\mathbf{G}^D)= \sum_{\mathbf{C}}\int \underbrace{p_{\theta}(\mathbf{Z}^{R})p_{\theta}(\mathbf{Z}^D|\mathbf{G}^{D})}_{\text{Latent embedding}} \notag \\
    &\underbrace{p(\mathbf{C}|\mathbf{Z}^D,\mathbf{Z}^R)}_{\text{Stochastic correlation matrix}} \underbrace{p_{\theta}(\mathbf{U}^D|\mathbf{C}, \mathbf{Z}^R)p(\mathbf{y}^{D}|\mathbf{U}^{D }, \mathbf{Z}^{D})}_{\text{Predictive distribution}} \notag \\
    &\underbrace{p(\mathbf{G}^D|\mathbf{Z}^D)}_{\text{Graph generator}}d\mathbf{Z}^Rd\mathbf{Z}^Dd\mathbf{U}^Dd\mathbf{G}.
\end{align}

\subsection{Graph Functional Neural Process With Learnable Rationales}


Our approach differs from the traditional FNP \citep{louizos2019functional} , which uses a randomly selected subset of the dataset (known as the reference set) to base its predictive distribution. This method lacks interpretability as it fails to provide summarized graph patterns for each class. In contrast, our approach involves learning a set of rationales for each class and basing the predictive distribution on these correlated rationales. The learned rationales represent the crucial substructures for each class.

\vspace{-0.5em}
\subsubsection{Learning rationales from probabilistic latent space.}
\vspace{-0.5em}
Learning the rationales directly in the graph domain is difficult due to its discrete nature, making it non-differentiable. Previous approaches, such as the one in \citet{you2018graph}'s work, have used reinforcement learning techniques to solve this issue through formulating the rationale generation as a Markov decision process. However, these methods often suffer from poor performance and instability during training \citep{wang2023gnninterpreter}. Our proposed solution is to learn a set of rationale embeddings in the continuous embedding space, which is easier to optimize and leads to improved performance.

Specifically, we first randomly initialize a set of vectors $\{\mathbf{s}_i^R\}_{i=1}^{|R|}$ for the rationales.
For a $K$-way classification problem, we assign $|R_k|$ rationales  ($|R|=\sum_{k=1}^K |R_k|$) to the $k$-th class, where $|R_k|$ is a hyper-parameter that we can tune during the training process. To capture the embedding uncertainty, we propose to further project $\mathbf{s}^R_i$ to a high-dimensional Gaussian distribution space:
\begin{equation}
  \mathbf{z}^R_i \sim \mathcal{N}( \text{MLP}(\mathbf{s}_i^R), \text{exp}(\text{MLP}(\mathbf{s}_i^R))).
\end{equation}
We denote the union of the parameters of $\{\mathbf{s}_i\}$ and the two MLPs as $\theta_{\rm r}$. The distribution of the rationale embeddings will be updated during training.

Each rationale embedding will encode one crucial predictive pattern for a specific class. We will base the predictive distribution of an input graph on such salient embeddings based on the correlations between the rationale embeddings and graph embeddings.  These rationale embeddings share a similar spirit with inducing points in stochastic variational Gaussian Process (SVGP, \citep{hensman2013gaussian}). However, we will enhance this by utilizing a generative model to decode the corresponding graph structures for model-level interpretability \ref{sec:decoder}.

Then we project the training graph into the same Gaussian distribution space through a GNN encoder. The GNN encoder computes node representations $\{\mathbf{h}_{i,v}|v\in \mathcal{V}_i\}$  as $ \{\mathbf{h}_{i,v} \} = \text{GNN}_e (G_i).$ The node vectors are aggregated to represent $G^D_i$ as a single vector $\mathbf{h}^D_i=\text{Aggregate} (\{\mathbf{h}_{i,v}\})$. Finally, the graph embedding follows: $\mathbf{z}_i^D \sim \mathcal{N}\left(\text{MLP}(\mathbf{h}_i^D), \text{exp}(\text{MLP}(\mathbf{h}_i^D))\right)$. We denote the union of the parameters of the GNN encoder and the two MLPs as $\theta_{e}$.

\subsubsection{Constructing stochastic correlation matrix $\mathbf{C}$}
The stochastic correlation matrix $\mathbf{C}$ models the relationship between the graphs and rationales in the embedding space, serving two important purposes: (1) it captures the uncertainty in the correlations for non-parametric uncertainty estimation, and (2) it identifies which rationale can be used to explain the corresponding prediction, thus improving the interpretability of the model.

Specifically, we first model the correlations in the latent embedding space using kernel similarity: $\kappa(\mathbf{z}^D_i, \mathbf{z}^R_{j})$, \eg,  we can use the radial basis function (RBF) kernel:
$\kappa(\mathbf{z}^D_i, \mathbf{z}^R_{j})={\rm exp}(-\gamma||\mathbf{z}^D_i-\mathbf{z}^R_j||)$.
Instead of directly using raw kernel similarity to parameterize, we further use Bernoulli sampling to generate a binary symmetric correlation matrix:
\begin{equation}
  \mathbf{C}_{i,j} \sim \text{Bern}(\mathbf{C}_{i,j}|\kappa(z^D_i,z^R_j)).
  \label{eq:c}
\end{equation}
This sampling process leads to sparse correlations for each sampled matrix and enjoys two benefits: (a) it can capture uncertainty from the data correlation perspective; (b) it can speed up model training by virtue of sparsity.

When $C_{i,j}=1$, it means that the $j$-th rationale $z^R_j$ is correlated with the $i$-th graph $G_i$. Through analyzing such a binary correlation matrix, we can identify which rationale represents a crucial pattern for
the input graph. 
\vspace{-0.7em}
\subsubsection{Constructing the predictive distribution}
With the binary correlation matrix, we summarize the information of the correlated rationales for each graph into the local rationale embedding $\mathbf{u}^D_i$:
{\small
\begin{equation}
  \mathbf{u}^D_{i} \sim \mathcal{N}(C_i\sum_{j:\mathbf{C}_{j,i}=1} \text{MLP}(\mathbf{z}^R_j),  \exp(C_i\sum_{j:\mathbf{C}_{j,i}=1} \text{MLP}(\mathbf{z}^R_j))),
  \label{eq:local}
\end{equation}
}
where  $C_i=\sum_{j}\mathbf{C}_{i,j}$ is for normalization.

As we can see, Equation~\ref{eq:local} encodes the inductive bias that predictions on points that are “far away,” i.e., have
very small probability of being connected to the rationales, will default to an uninformative standard Gaussian prior. This is similar to the behavior that
Gaussian processes (GPs) with RBF kernels exhibit.

The local rationale embeddings are derived solely from the rationale embedding set and may not capture novel information present in the graphs. To address this issue, we include the graph embedding $\mathbf{z}_i$ in the final prediction. This allows the neural network to extrapolate beyond the distribution of the rationale embeddings, which is important when the unseen test graphs contain novel predictive patterns that are not present in the learned rationale set. Therefore, we concatenate the graph embedding and local rationale embedding into a single vector and obtain the final predictive distribution:
\begin{equation}
  y_i = \text{MLP}(\text{concat}(\mathbf{z}_i, \mathbf{u}_i)).
  \label{eq:prediction}
\end{equation}
We denote the parameters of the MLPs in Eq.~\ref{eq:local} and Eq.~\ref{eq:prediction} as $\theta_{\rm cls}$.

\vspace{-0.7em}
\subsection{Graph Structure Decoder}
\label{sec:decoder}
\vspace{-0.5em}
The goal of learning rationales in the embedding space is to understand the graph structure that drives the model's predictions. To achieve this, we design a graph generator that can decode the rationales from the embedding space and produce graph structures.


To ensure that our decoded rationale structures contain critical patterns for classification, we propose a variant of the GraphRNN model \citep{you2018graphrnn}. This variant incorporates latent embedding information into the graph generation process. The graph is generated in breadth-first order, with each step involving the generation of a node and its edges, taking into account all previously generated nodes. Unlike standard GraphRNN models, which are only used for random graph generation, our proposed model ensures that the decoded rationales retain the critical patterns necessary for accurate classification.

Specifically, in $t$-th step, the decoder first runs a decoder GNN over current graph $G_t$ to compute node representations:$
  \{ \mathbf{h}_v^t\} = \text{GNN}_d(G_t).$ The current graph $G_t$ is represented as an aggregation of its node vectors $\mathbf{h}_{G_t}=\text{Aggregate}(\{\mathbf{h}_v^t\})$. With the current graph representation,  we predict the probability of the node type of $v_{t}$ as:
\begin{equation}
  \mathbf{p}_{v_{t}}=\text{softmax}(\text{MLP}(\mathbf{h}^t_{v_t},\mathbf{h}_{G_t},\mathbf{z})).
  \label{eq:node}
\end{equation}
Note that it is easy to generalize to continuous node features by assuming $\mathbf{p}_{v_t}$ follows a Gaussian distribution.  To fully capture edge dependencies, we predict the edge type between $v_t$ and all the previously generated node $\{v_j \}_{j=1}^{t-1}$ sequentially and update the representation of $v_t$ when a new edge is added to $G_t$. In the $j$-th step, we predict the edge type between $v_t$ and $v_j$ as:
\begin{equation}
  \mathbf{p}_{e_{v_t, v_j}} = \text{softmax}(\text{MLP}(\mathbf{h}_{v_t}^{t,j}, \mathbf{h}^t_{v_j},\mathbf{h}_{G_t},\mathbf{z})),
  \label{eq:edge}
\end{equation}
where $\mathbf{h}^{t,j}_{v_t}$ is the new representation of $v_t$ after the edge $e_{v_t, v_j}$ has been added.
We denote the parameters of the graph decoder as $\theta_{\rm d}$. 

As seen from Eq.~\ref{eq:node} and Eq.~\ref{eq:edge}, the generation procedure of nodes and edges both depend on the latent graph embedding $\mathbf{z}$.
This latent embedding holds crucial information about graph properties, and the learned rationale embeddings are salient points within the same space. Hence, the decoded rationale structure should reflect the critical subgraph patterns of its corresponding class.

\begin{table*}[]
  \small
  \centering
  \fontsize{7.5}{9.5}\selectfont
  \setlength{\tabcolsep}{0.4em}
  \caption{ECE and predictive performance on all the five datasets. We report the average performance and standard deviation for 5 random initializations. For predictive performance, Graph-SST2 and MUTAG use accuracy as the metric, while BBBP and BACE use AU-ROC.}
  \vspace{2ex}
  \resizebox{1\textwidth}{!}{%
  \begin{tabular}{@{}llccccc|ccccc@{}}
    \toprule
    &              & \multicolumn{4}{c}{ECE} &  & \multicolumn{4}{c}{Accuracy/AU-ROC} &  \\ \cmidrule(l){3-12}
    \multicolumn{1}{l}{Backbone} & \multicolumn{1}{l}{Model} & Graph-SST2 & BBBP & BACE & MUTAG & Github & Graph-SST2 & BBBP & BACE & MUTAG & Github  \\ \midrule
    \multicolumn{1}{l|}{\multirow{9}{*}{GCN}} & Vanilla      &$15.07$\std{0.60}      & $18.30$\std{1.56}       & $19.52$\std{2.87}      & $3.54$\std{1.05}     & $6.73$\std{0.99} &  $82.83$\std{0.58}        &  $69.34$\std{1.29}        &  $77.52$\std{0.66}       &   $79.38$\std{0.87}     &  $61.26$\std{0.39}\\
    \multicolumn{1}{l|}{}                     & DropEdge     & $13.89$\std{3.48}      & $16.78$\std{2.11}      & $16.23$\std{3.33}   &  $4.57$\std{0.79}   & $7.15$\std{0.67} &      $82.89$\std{1.31}    &  $70.02$\std{0.69}        & $76.07$\std{2.40}        &  $76.50$\std{1.63}      &  $61.09$\std{0.64}\\
    \multicolumn{1}{l|}{}                     & MC-Dropout   & $13.82$\std{0.73}     & $15.53$\std{1.38}      &  $18.76$\std{2.63}    &   $3.49$\std{0.62}   & $6.28$\std{0.69}  &    $83.21$\std{0.64}      &  $69.62$\std{1.13}      & $78.40$\std{0.65}    &  $78.34$\std{0.81}  & $63.29$\std{0.84} \\
    \multicolumn{1}{l|}{}                     & SGLD         &$10.71$\std{0.59}       &  $15.66$\std{2.71}     & $14.47$\std{3.61}     & $5.76$\std{0.74}     & $3.96$\std{1.84} &   $82.01$\std{0.62}       &   $67.83$\std{2.33}       & $76.74$\std{1.87}      &    $77.89$\std{0.59}      &  $59.49$\std{0.21}\\
    \multicolumn{1}{l|}{}                     & BBP          &$14.57$\std{0.81}       &  $14.04$\std{1.30}     & $13.94$\std{1.22}     & $7.26$\std{1.24}     & $4.35$\std{0.93} & $82.72$\std{0.67}        &   $69.54$\std{2.26}       &   $75.34$\std{1.37}      &   $76.82$\std{0.59}     &  $61.06$\std{0.62}\\
    \multicolumn{1}{l|}{}                     & Graph-GP     &   $12.52$\std{0.76}   &  $14.36$\std{1.98}     & $14.31$\std{2.87}     & $4.96$\std{1.32}    & $5.32$\std{0.87} &    $81.01$\std{0.87}      &  $68.98$\std{1.76}        &   $76.96$\std{0.97}      & $78.16$\std{1.03}       & $60.75$\std{1.26} \\
    \multicolumn{1}{l|}{}                     & GSAT         &  $16.51$\std{2.11}     &  $37.33$\std{0.71}    & $31.32$\std{1.66}    &  $8.64$\std{1.86}   & $4.04$\std{0.50}  &  $81.09$\std{1.04}        &  $71.55$\std{1.56}         &  $77.91$\std{0.78}      &   $78.89$\std{0.58}     & $60.42$\std{0.11} \\
    \multicolumn{1}{l|}{}                     & DeepEnsemble &$9.81$\std{NA}       & $15.00$\std{NA}      &  $16.04$\std{NA}    &  $3.22$\std{NA}   & $6.21$\std{0.15} &    $84.71$\std{NA}      &  $69.75$\std{NA}        &  $78.17$\std{NA}       &  $79.03$\std{NA}      & $61.97$\std{0.65}  \\
    \rowcolor[HTML]{EFEFEF}
    \multicolumn{1}{l|}{}                     & \ours        & $\bm{9.02}$\std{0.58} & $\bm{10.95}$\std{0.92} & $\bm{9.11}$\std{2.96} & $3.82$\std{1.07} & $\bm{3.25}$\std{0.42} & $83.08$\std{0.26} & $70.25$\std{0.70} & $80.36$\std{0.89} & $79.80$\std{0.71} & $61.34$\std{0.85} \\ \midrule
    \multicolumn{1}{l|}{\multirow{9}{*}{GAT}} & Vanilla      &$13.73$\std{0.60}       &   $23.40$\std{5.30}   & $17.84$\std{2.45}    &  $4.18$\std{0.71}   & $5.72$\std{1.05}& $83.63$\std{0.82}  & $69.82$\std{2.17} &     $77.40$\std{3.85}                &    $72.88$\std{1.34}    &  $62.19$\std{1.75}\\
    \multicolumn{1}{l|}{}                     & DropEdge     &$10.33$\std{0.62}      &  $20.34$\std{1.68}     &  $15.39$\std{3.07}    &  $5.32$\std{1.67}  & $9.65$\std{1.03}  &   $83.76$\std{0.60}      &  $66.85$\std{0.74}       &    $80.10$\std{5.76}     &  $70.04$\std{2.61}       & $61.67$\std{0.89} \\
    \multicolumn{1}{l|}{}                     & MC-Dropout   &  $12.32$\std{0.73}    & $19.85$\std{3.93 }      &  $15.20$\std{2.85}    &  $\bm{3.74}$\std{0.65}    & $5.03$\std{0.78}  &    $83.46$\std{0.42}     &  $70.03$\std{1.98}       &  $77.82$\std{3.71}     &   $72.72$\std{0.63}     &  $64.26$\std{0.75}\\
    \multicolumn{1}{l|}{}                     & SGLD         &$10.28$\std{0.31}       &  $17.66$\std{1.95}     & $20.81$\std{4.94}    &  $4.66$\std{0.57}   & $6.21$\std{1.48}  &   $83.67$\std{0.41}     & $67.87$\std{2.34}         &   $74.25$\std{3.67}      &   $72.02$\std{1.33}     & $60.75$\std{0.57} \\
    \multicolumn{1}{l|}{}                     & BBP          &$9.26$\std{1.94}       & $16.45$\std{1.47}      &  $14.24$\std{6.25}    &  $6.09$\std{1.20}   &$4.95$\std{0.89}   &     $80.82$\std{0.59}    &  $68.89$\std{1.62}        & $74.78$\std{5.63}        &    $72.63$\std{1.75}    & $60.56$\std{1.08} \\
    \multicolumn{1}{l|}{}                     & Graph-GP     &  $11.96$\std{1.86}    & $18.25$\std{2.01}      & $15.86$\std{1.75}     &  $4.75$\std{1.54}    & $4.32$\std{0.86}  &      $82.75$\std{1.86}   &   $68.78$\std{2.01}      & $76.41$\std{2.76}       &  $72.26$\std{1.59}      & $61.94$\std{1.03} \\
    \multicolumn{1}{l|}{}                     & GSAT         &   $13.54$\std{2.04}   & $34.17$\std{0.85}     &  $32.08$\std{0.97}  &  $7.91$\std{2.10}   & $3.90$\std{2.79} &     $81.98$\std{1.87}     &    $68.65$\std{1.06}       &  $72.23$\std{1.54}      &  $72.02$\std{1.65}      & $62.22$\std{1.50}  \\
    \multicolumn{1}{l|}{}                     & DeepEnsemble &  $8.43$\std{NA}    & $16.80$\std{NA}      & $17.73$\std{NA}     &  $4.35$\std{NA}   & $4.92$\std{0.29} &  $85.69$\std{NA}      &   $71.98$\std{NA}      &   $79.20$\std{NA}      & $73.27$\std{NA}& $63.89$\std{0.72} \\
    \rowcolor[HTML]{EFEFEF}
    \multicolumn{1}{l|}{}                     & \ours        &$\bm{7.89}$\std{1.61}       & $\bm{15.76}$\std{2.34}      &$\bm{13.58}$\std{3.43}      &  $3.99$\std{0.23}  &$\bm{3.74}$\std{0.85}   &   $83.45$\std{1.56}      &  $70.07$\std{0.63}        & $79.13$\std{1.96}       &    $74.19$\std{0.33}    & $62.75$\std{1.03} \\ \bottomrule
  \end{tabular}
  }
  \label{table:ece}
  \vspace{-0.5em}
\end{table*}

\subsection{Model Training}
We now describe how to learn the model parameters during training. The rationales and training graphs are intertwined in the same space, making direct optimization of the overall data likelihood (Eq. 1) difficult. To solve this, we employ an alternating optimization strategy, similar to the Expectation-Maximization (EM) approach. 

\paragraph{E-step}
In this step, we aim to update the rationale embeddings and graph decoder given a fixed graph encoder and classifier. Canceling out the constant terms, we have the following loss:
\begin{align}
  &\text{argmin}_{\theta_{\rm d}, ¥\theta_{\rm r}}\mathcal{L}_{\text{E}} \notag \\
  &= \underbrace{-\mathrm{E}_{p_{\theta}(\mathbf{Z}^R)p_{\theta}(\mathbf{Z}^D, \mathbf{U}^D, \mathbf{C}|\mathbf{G}^D, \mathbf{Z}^R)}  \log p(y^R|\mathbf{Z}^R, \mathbf{U}^R)}_{\text{Classification loss of training data}}\nonumber \\
  &-\underbrace{\mathrm{E}_{p_{\theta}(\mathbf{Z}^D|\mathbf{G}^D)} \log p_{\theta_d}(\mathbf{G}^D|\mathbf{Z}^D)}_{\text{Generation loss}} \nonumber\\
  &-\underbrace{\mathrm{E}_{p(\mathbf{Z}^R)p(\mathbf{U}^R|\mathbf{Z}^R)}\log p_{\theta}(\mathbf{y}^{D}|\mathbf{Z}^D,\mathbf{U}^D)}_{\text {Classification loss of rationales}}
    \label{eq:estep}
\end{align}

Note that, in Eq.~\ref{eq:estep}, we also add the classification loss of the rationales. For the rationales, as their rationale embedding already represents crucial predictive patterns, we directly project them into the local rationale embedding space with the MLP used in Eq.~7, \ie $\mathbf{u}^R_{i}\sim \mathcal{N}( \text{MLP}(\mathbf{z}_i^R), \text{exp}(\text{MLP}(\mathbf{z}_i^R))$.

\paragraph{M-step}
In this step, we optimize the parameters of the probabilistic encoder and classifier by fixing the distributions of the rationale embeddings and graph decoder. Directly maximizing the data likelihood is intractable and we choose to use the amortized variational inference. Following the work of \citet{louizos2019functional}, we assume that the variational posterior distribution $q_{\phi}(\mathbf{U}^D,\mathbf{C},\mathbf{Z}^D|\mathbf{G}^D)$
factorizes as $p_{\theta}(\mathbf{Z}^D|\mathbf{G}^D)p(\mathbf{C}|\mathbf{Z}^R,\mathbf{Z}^D)q_{\phi}(\mathbf{U}^D|\mathbf{G}^D)$. This leads to
the following loss:
{\small
\begin{align}
  &\text{argmin}_{\theta_{\rm e}, \theta_{\rm cls} , \phi}\mathcal{L}_{\text{M}}  \nonumber \\
  &=
    \underbrace{-\mathrm{E}_{p_{\theta}(\mathbf{Z}^R)q_{\phi}(\mathbf{U}^D|\mathbf{G}^D)}\log  p_{\theta}(\mathbf{y}^{D}|\mathbf{Z}^D,\mathbf{U}^D)}_{\text{Classification loss of training data}} \nonumber\\
  &\underbrace{-{\rm E}_{p_{\theta}(\mathbf{Z}^R) p_{\theta}( \mathbf{C}, \mathbf{Z}^D|\mathbf{G}^D, \mathbf{Z}^R)q(\mathbf{U}^D|\mathbf{G}^D)}\log \frac{ p_{\theta}(\mathbf{U}^D|\mathbf{C}, \mathbf{Z}^R,\mathbf{Z}^D)}{ q_{\phi}(\mathbf{U}^{D}|\mathbf{G}^D)}}_{\text{Prior regularization}} \nonumber\\
  &\underbrace{- \mathrm{E}_{p_{\theta}(\mathbf{Z}^R)p(\mathbf{U}^R|\mathbf{Z}^R)}p_{\theta}(y^R|\mathbf{Z}^R, \mathbf{U}^R)}_{\text{Classification loss of rationales}}
\end{align}
}

As sampling from the Bernoulli distribution in Equation \ref{eq:c} leads to
discrete correlated data points, we make use of the Gumbel softmax trick \citep{jang2016categorical} to make the model differentiable. 

\vspace{-0.7em}
\section{Experiments}
\vspace{-0.7em}

In this section, we evaluate the empirical performance of our model \ours.
Our experiments are designed to answer the following questions: Q1: Can \ours provide calibrated predictive uncertainty? and Q2: How is the predictive power of  the learned rationales? Q3: Does the decoded rationale graph structure provide meaningful explanations? Q4: How important are the different components of \ours (Appendix~\ref{apd:abs})?

\subsection{Q1: Can \ours provide calibrated predictive uncertainty?}
We first examine if \ours can simultaneously provide calibrated uncertainty estimates and maintain high predictive accuracy.

\begin{table*}[t]
  \small
  \fontsize{7.5}{9.5}\selectfont
  \caption{RF1 of all the methods. $K$ is the number of neighbors in the KNN classifier. We report the average performance and
    standard deviation for 5 random seeds. Our method consistently outperforms all the baselines in RF1 across different datasets
    and GNN architectures. XGNN cannot be applied on Graph-SST2 because it can only generate graphs with discrete node features.}
  \vspace{2ex}
  \setlength{\tabcolsep}{0.4em}
  \resizebox{1\textwidth}{!}{%
  \begin{tabular}{llccccl|ccccl}
    \hline
    &              & \multicolumn{5}{c|}{$K=1$} & \multicolumn{5}{c}{$K=3$} \\ \cline{3-12}
    \multicolumn{1}{c}{Backbone} & Model & Graph-SST2 & BBBP & BACE & MUTAG & Github & Graph-SST2 & BBBP & BACE & MUTAG & Github \\ \hline
    \multicolumn{1}{l|}{\multirow{7}{*}{GCN}} & GNNExplainer &  $80.45$\std{0.30}  & $63.93$\std{2.66}   & $67.38$\std{0.59}   &  $64.36$\std{0.87} &  $54.37$\std{0.16}  & $80.46$\std{0.17}   & $64.98$\std{1.49}   & $68.61$\std{2.17}  & $65.46$\std{1.64}  &   $58.78$\std{0.07}  \\
    \multicolumn{1}{l|}{}                     & PGExplainer  & $80.16$\std{0.13}   &  $59.75$\std{2.46}  & $61.53$\std{3.64}   & $71.52$\std{0.61}   &  $45.25$\std{0.54}  & $80.69$\std{0.17}   &  $63.32$\std{0.30}  & $66.70$\std{0.99}  & $74.79$\std{2.57}  &  $52.79$\std{0.36}  \\
    \multicolumn{1}{l|}{}                     & XGNN         & NA & $56.23$\std{4.25} & $53.62$\std{3.72}    & $68.67$\std{3.01}  & $51.92$\std{1.92}  & NA   & $52.28$\std{6.29}    &  $58.96$\std{2.82} & $65.86$\std{1.96}   & $50.29$\std{1.85}     \\
    \multicolumn{1}{l|}{}                     & GstarX       & $81.60$\std{0.14}   & $58.55$\std{2.40}   &  $65.84$\std{1.62}  & $69.47$\std{2.84}  &   $56.97$\std{1.21} & $81.64$\std{0.17}   &  $61.00$\std{6.40}  & $65.00$\std{0.43}   & $68.32$\std{1.17}  &  $60.12$\std{2.76}   \\
    \multicolumn{1}{l|}{}                     & GSAT         &  $79.13$\std{1.35}  & $62.62$\std{2.46}   & $63.76$\std{5.96}   & $53.96$\std{14.59}   & $58.19$\std{2.96}  &   $78.45$\std{2.73}   & $64.26$\std{4.65}  & $64.98$\std{1.27}  & $65.24$\std{7.35}   &  $56.20$\std{2.49}   \\
    \multicolumn{1}{l|}{}                     & SubgraphX    &  $82.56$\std{0.09}  & $63.70$\std{0.49}   & $68.24$\std{2.45}   & $\bm{75.19}$\std{0.73}  &  $57.28$\std{0.86}  &   $82.64$\std{0.19} &  $64.62$\std{0.90}  & $44.35$\std{4.38}  & $68.17$\std{7.78}  &  $60.93$\std{1.29} \\
    \rowcolor[HTML]{EFEFEF}
    \multicolumn{1}{l|}{}                     & \ours        & $\bm{85.85}$\std{0.67} & $\bm{67.42}$\std{1.85} &  $\bm{71.52}$\std{2.53}  & $\bm{74.47}$\std{0.48}  & $\bm{64.62}$\std{1.07}   & $\bm{85.84}$\std{0.64}   & $\bm{67.74}$\std{2.20}    & $\bm{71.52}$\std{2.52}  & $\bm{74.42}$\std{0.45}   &   $\bm{64.46}$\std{1.38}  \\ \hline
    
    \multicolumn{1}{l|}{\multirow{7}{*}{GAT}} & GNNExplainer & $77.23$\std{1.42}   & $60.52$\std{5.52}   & $47.75$\std{5.57}   & $59.20$\std{1.42}  &  $51.71$\std{0.80}  & $78.49$\std{1.76}   &  $48.76$\std{14.81}  & $44.86$\std{3.75}  &  $66.28$\std{2.19} & $58.40$\std{0.28}    \\
    \multicolumn{1}{l|}{}                     & PGExplainer  &  $79.59$\std{1.87}  & $55.38$\std{6.61}   & $54.56$\std{3.65}   &  $58.81$\std{2.34} &  $48.76$\std{1.45}  &  $80.22$\std{2.54}  & $66.44$\std{2.37}   & $48.31$\std{10.88}   & $59.51$\std{1.74}  &  $49.09$\std{1.21}  \\
    \multicolumn{1}{l|}{}                     & XGNN         & NA &  $56.26$\std{2.76}    & $49.21$\std{3.28}    & $64.54$\std{3.06}  & $52.86$\std{3.76}  & NA  & $50.75$\std{5.06}  & $60.76$\std{10.23}   & $62.07$\std{4.28}  & $50.20$\std{6.08}   \\
    \multicolumn{1}{l|}{}                     & GstarX       &  $79.82$\std{2.21}  & $58.06$\std{2.02}   & $38.28$\std{6.52}   & $61.14$\std{3.90}  & $56.12$\std{2.18}   & $80.48$\std{1.42}   & $63.24$\std{1.09}   &  $55.26$\std{2.84} &  $66.87$\std{0.52} &  $59.21$\std{1.46}  \\
    \multicolumn{1}{l|}{}                     & GSAT         & $77.43$\std{2.09}   & $61.23$\std{24.10}   & $62.95$\std{6.20}   & $63.72$\std{3.47} &  $55.25$\std{3.52} &   $78.04$\std{1.87}   & $63.37$\std{18.70}   & $63.67$\std{5.86}   & $63.96$\std{4.32}   &  $59.86$\std{2.76}  \\
    \multicolumn{1}{l|}{}                     & SubgraphX    &  $82.29$\std{1.56}  & $64.16$\std{2.99}   & $58.46$\std{1.47}   & $66.73$\std{1.70}  & $54.26$\std{1.72}    & $81.96$\std{1.79}   &  $64.85$\std{1.91}  & $57.91$\std{2.26}   &  $63.83$\std{0.35} & $56.34$\std{1.29}   \\
    \rowcolor[HTML]{EFEFEF}
    \multicolumn{1}{l|}{}                     & \ours        &  $\bm{85.67}$\std{2.21}  & $\bm{67.99}$\std{3.00}   & $\bm{70.00}$\std{6.28}   & $\bm{68.75}$\std{0.27}  &  $\bm{61.38}$\std{1.26}  & $\bm{85.66}$\std{1.90}   & $\bm{68.23}$\std{3.01}   & $\bm{70.57}$\std{5.85}  & $\bm{69.08}$\std{0.32}  &  $\bm{63.27}$\std{1.19}  \\ \hline
  \end{tabular}
  }
  \label{table:rationale}
  \vspace{-1em}
\end{table*}

\label{subsection:ece}
\vspace{-0.8em}
\subsubsection{Experimental Setup}
We use
\textbf{Expected calibration error (ECE)} to evaluate  the model calibration \citep{guo2017calibration, naeini2015obtaining}.   

We use the following five graph classification datasets:
\noindent$\bullet$\textbf{BBBP} is designed for the modeling and prediction of barrier permeability. This dataset includes binary labels for 2039 compounds on their permeability properties. $\bullet$\textbf{BACE} is a collection of 1522 compounds with their binary labels for a set of inhibitors of human $\beta$-secretase 1. $\bullet$\textbf{Graph-SST2} is a sentiment analysis dataset, where each text sequence in SST2 is converted to a graph. Following the splits in the study of \citet{wu2022discovering}, this dataset contains degree shifts during test. $\bullet$\textbf{MUTAG} consists of 4,337 molecule graphs. Each graph is assigned to one of 2 classes based on its
mutagenic effect. $\bullet$\textbf{Github} has 12,725 social networks and the task is to predict whether a  network belongs to web or machine learning developers.

\vspace{-0.7em}
\subsubsection{Baselines}
\vspace{-0.5em}

We consider the following baselines:
\textbf{Vanilla}\citep{hendrycks2016baseline} directly uses the model's softmax score of the predicted category as the uncertainty estimate.  \textbf{Monte Carlo dropout (MC-dropout)} \citep{gal2016dropout}
applies dropout at test  time for multiple times and then average the outputs.  \textbf{Stochastic Gradient Langevin Dynamics (SGLD)} \citep{welling2011bayesian} is the most popular Markov Chain Monte Carlo (MCMC) based Bayesian neural network (BNN). \textbf{Bayes by Backprop (BBP)} \citep{blundell2015weight}  fits a variational approximation to the true posterior of BNN with a fully factorised Gaussian assumption. \citep{kingma2015variational} as a variance reduction technique. \textbf{GSAT} \citep{miao2022interpretable} learns a stochastic attention to select task relevant subgraphs.
\textbf{DropEdge} \citep{Rong2020DropEdge:} randomly removes a certain number of edges from the input graph during training. \textbf{Graph-GP} \citep{ng2018bayesian} adopts GNN as the feature extractor and replaces the linear classifier layer with a Gaussian process. \textbf{DeepEnsemble}\citep{lakshminarayanan2017simple} trains multiple GNNs with different initializations and aggregates their predictions at inference time. We use 5 as the ensemble size.

\begin{figure*}[!t]
  \centering
  \includegraphics[width=0.9\linewidth, height=4.5cm]{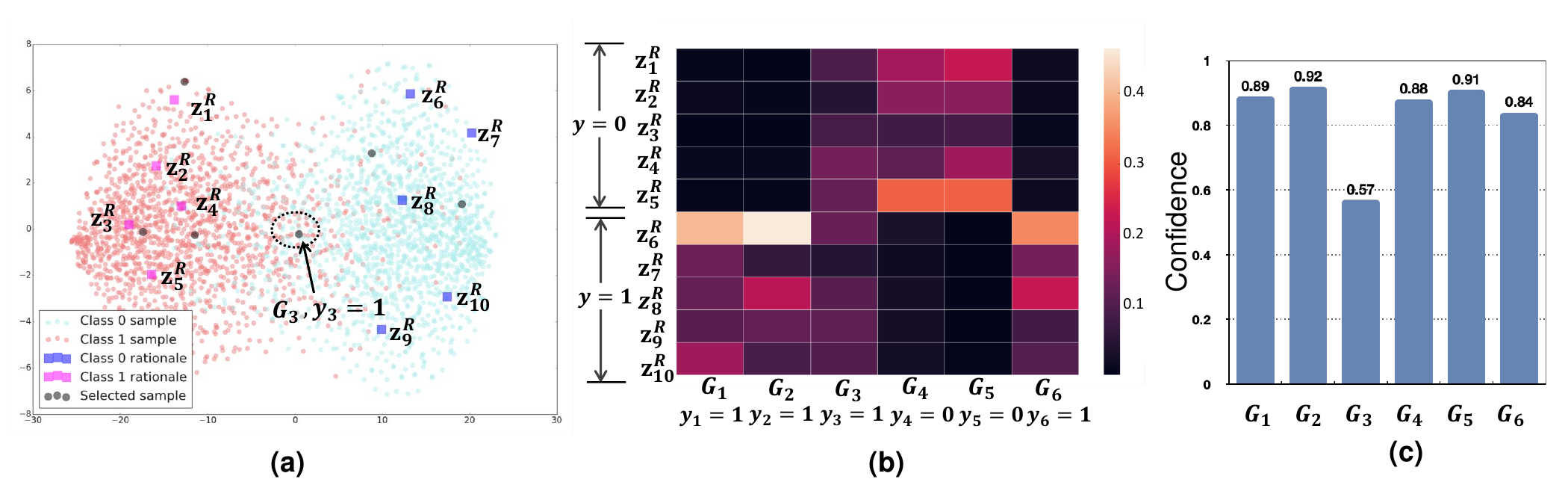}
  \caption{(a) visualizes the sampled rationale embeddings and graph embeddings. (b) plots the averaged stochastic correlations matrix for the selected six samples. (c) shows the predictive confidence of sample 1-3.}
  \label{fig:embedding}
\end{figure*}
\vspace{-0.7em}
\subsubsection{Main Results}
\vspace{-0.5em}
Table~\ref{table:ece} reports the ECE and predictive performance of all the methods. \ours outperforms all the baselines on all the datasets except for MUTAG. On MUTAG, the Vanilla GNN has already achieved a good calibration score, and thus the room for further improvement is small. We can also see that the ECE of some baselines, such as BBP, are even worse than the Vanilla GNN on this dataset. \ours can improve ECE up to $4.83\%$  with GCN as the backbone and $1.62\%$ with GAT as the backbone compared with the strongest baseline. The consistent improvement among GCN and GAT verifies that \ours is a general method that can be compatible with existing GNN architecture.

Though with better calibration scores, existing Bayesian uncertainty estimation methods often suffer from decreased predictive performance. For example, on the BACE dataset, the predictive accuracy of SGLD drops by $3.15\%$ compared with Vanilla GAT. However, \ours can consistently boost the performance of Vanilla GNN. For example, on BACE, we improve the AU-ROC by $2.84\%$ compared with Vanilla GCN.
This is because \ours quantifies uncertainty from the functional space and does not need to specify the uninterpretable prior distribution of model parameters as existing BNNs.

\begin{figure*}
  \centering
  \includegraphics[width=0.9\textwidth]{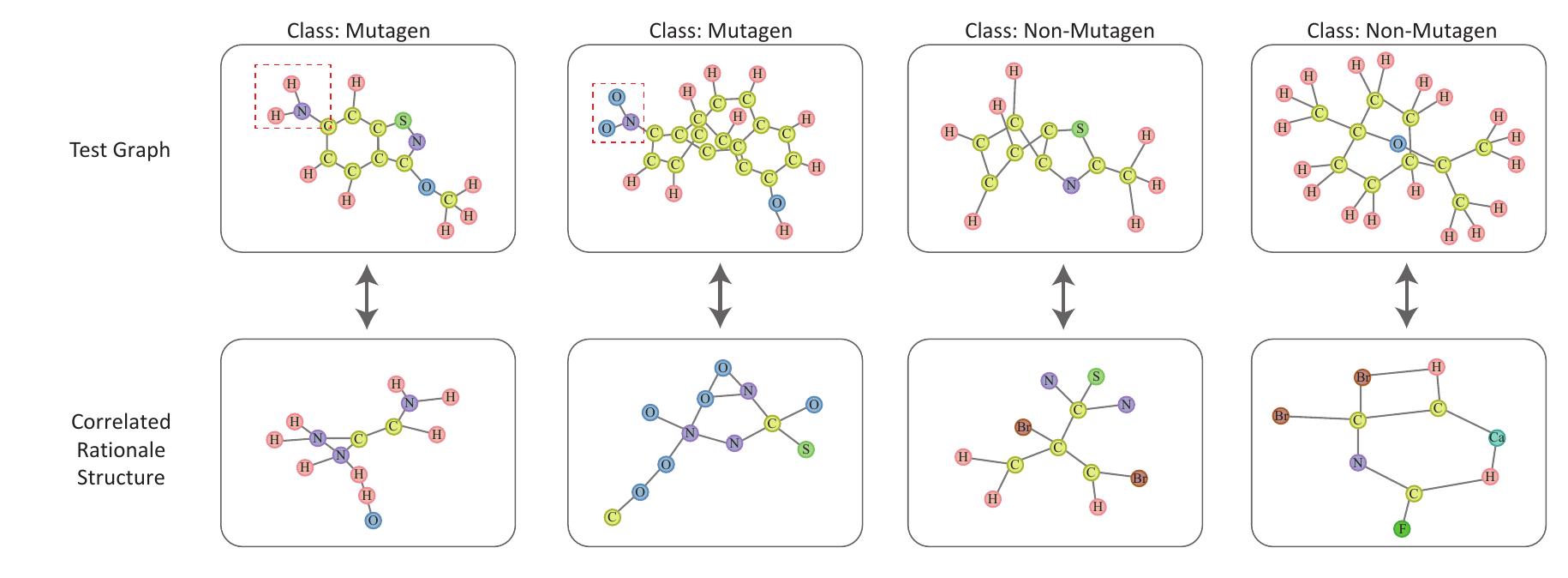}
  \caption{The decoded graph structure of the correlated rationale. The first row is the input test graphs, while the second shows the corresponding correlated model-level rationale structures. The rationales for the mutagen class contain repeated patterns of NH2 and NO2. While the rationales for the non-mutagen class have no particular patterns. }
  \vspace{-2ex}
  \label{fig:structure}
\end{figure*}

\vspace{-0.7em}
\subsection{Q2:  How is the predictive power of the learned rationales?}
\vspace{-0.5em}
We have verified that \ours can provide accurate and calibrated predictions. In this subsection, we further study whether \ours can learn high-quality rationales.
\vspace{-0.5em}
\subsubsection{Experimental Setup}
\vspace{-0.5em}

Ideally, the rationale set should contain critical class information and achieve high predictive performance. We measure the predictive power of the discovered rationales by using a KNN classifier. To apply the KNN classifier, we compute the cosine distance between the test graph and the rationales in the embedding space. We refer to this metric as
"\textbf{Rationale F1 (RF1)}".
In the experiments, we set $K$ to 1 and 3.

We compare with  the following baselines:
\textbf{GNNExplainer} \citep{ying2019gnnexplainer}  learns sample-level rationales by minimizing the mutual information between
the graph prediction and distribution of possible subgraphs.
\textbf{PGExplainer} \citep{luo2020parameterized} uses a deep neural network to parameterize the generative process of the underlying graph structure as edge distributions, where the explanatory graph is sampled. \textbf{SubgraphX} \citep{yuan2020xgnn} searches representative subgraphs using Monte Carlo tree search with Shapley values as the importance score. \textbf{GStarX} \citep{zhang2022explaining} proposes to use cooperative game theory to extract important subgraphs. \textbf{GSAT} \citep{miao2022interpretable} learns a stochastic attention weight to select task-relevant subgraphs.
\noindent$\bullet$ \textbf{XGNN} \citep{yuan2020xgnn} trains a deep reinforcement learning model to generate the model-level explanation graphs.

All the baselines except for XGNN are originally designed to obtain sample-level rationales. To obtain model-level rationales, we first apply them to the training data and then perform centroid clustering for each class. The centroid of each cluster is used as the model-level rationale for each class.

\vspace{-0.5em}
\subsubsection{Experimental Results}
Table~\ref{table:rationale} reports the RF1 of all the methods. As shown, our method consistently outperforms other methods on all the datasets for both $K=1$ and $K=3$.  The stable improvements of our
method across GCN and GGNN demonstrates that our proposed
the framework is general and can be applied to any GNN architectures.

Fig.~\ref{fig:embedding} shows the sampled graph and rationale embeddings. As shown in Fig.~\ref{fig:embedding}(a), the rationale embeddings for each class are evenly distributed within their respective class cluster, demonstrating that each rationale encodes a critical predictive pattern. Fig.~\ref{fig:structure}(b) depicts the average stochastic correlation matrix for six selected samples. It can be seen that the most correlated rationale for each sample is from the same class. The average values for rationales from other classes are close to zero for all samples except for sample 3, which is located near the boundary of two classes and has similar distances to the closest rationales of both classes, resulting in high uncertainty in model predictions. On the other hand, other samples are closely positioned to the rationales within their class cluster, leading to more confident predictions by the model.

\vspace{-0.5em}
\subsection{Q3: Does the decoded graph structure of the rationale provide meaningful explanations?}

In this subsection, we study whether the decoded graph structure of the correlated rationale can provide meaningful explanations for the model prediction.

We start by visualizing the synthetic BA2Motifs dataset, which consists of graphs built on an arabasi-Albert (BA) structure. Half of the graphs are attached with a house motif, and the other half with a five-node cycle motif. As seen in Fig.~\ref{fig:intro}(b), the decoded graph structure of the selected rationale is a five-node cycle, aligning with the data generation process.

We further validate our method on the real-world MUTAG dataset. As reported by \cite{doi:10.1021/jm00106a046}, the presence of chemical groups NH2 and NO2 in molecules is associated with mutagenicity. Fig.~\ref{fig:structure} shows the results on this dataset. The first two graphs in the first row are from the mutagen class and have either one NH2 or one NO2 group. The decoded rationales of these graphs, shown in the second row, reveal repeated patterns of NH2 and NO2, respectively. Conversely, the decoded rationale structures for the non-mutagen class do not display any specific patterns, which aligns with our prior knowledge.

The above results demonstrate that our decoded rationale structures contain crucial substructures for each class. These model-level rationales offer practitioners valuable insights into domain knowledge and help confirm if the model's workings align with real-world rules.
\vspace{-1em}

\vspace{-0.7em}
\section{Related Works}
\vspace{-0.7em}

\noindent\textbf{Uncertainty quantification on GNNs}. \citet{ng2018bayesian} propose 
Graph Gaussian Process (Graph-GP), which stacks a 
Bayesian linear model on the feature representation of GNNs. \citet{liu2020uncertainty}
further, extend Graph-GP by considering the uncertainty in the input graph.
 \citet{zhao2020uncertainty} proposes
subjective GNN considering multi-source uncertainty.
\citet{10.5555/3524938.3525321} adaptively drops some edges during training and shows that it approximates the variational inference of BNNs. 
\citet{stadler2021graph} generalize the Posterior Network \citep{10.5555/3495724.3495839} to graph data. Along another line, \citet{jian, wang2021confident} propose two post-hoc methods to improve GNN's calibration.
However, these works are all designed for node classification and semi-supervised learning, while our work focuses on the graph-level property classification.  Localized Neural Kernel (LNK, \citep{wollschlager2023uncertainty}) primarily concentrates on energy prediction and a molecule’s energy is additive concerning its atoms. Therefore, it initially defines a Gaussian process at the atom level, followed by summing up the energy predictions from each node for the final regression output. In contrast, our paper tackles graph-level classification tasks. We opt to aggregate node-level representations into a global one for classification purposes.

\noindent\textbf{Explaining GNNs}.
There have been a large number of works in generating 
 sample-level rationales for GNNs. These methods aim to provide the salient patterns for a specific input graph. According to \citet{yuan2022explainability}'s work, most existing works can be categorized into four directions \citep{wu2022survey}: gradient-based methods \citep{baldassarre2019explainability, pope2019explainability}, pertubation-based methods \citep{ying2019gnnexplainer, luo2020parameterized, zhang2022explaining, funke2021hard, schlichtkrull2021interpreting}, decomposition methods \citep{baldassarre2019explainability, schwarzenberg2019layerwise, schnake2021higher, wang2021causal} and surrogate methods \citep{huang2022graphlime, vu2020pgm, zhang2021relex}.
 In contrast, model-level rationale generation remains underexplored. It aims to explain the overall behavior of the model and is not specific to any particular graph. XGNN \citep{yuan2020xgnn} proposes to formulate the model-level rationale generation as a Markov decision process and use reinforcement learning to for optimization.  DAG-Explainer \citep{lv2023data} proposes a data-aware global explainer based on randomized greedy algorithm.  
 Recently, \citet{wang2023gnninterpreter} proposes a numerical optimization method to obtain the explanation graph via continuous relaxation. Both of these two works are post-hoc methods while we develop an inherently interpretable model. Existing work \citep{miao2022interpretable} has shown that the post-hoc interpretability methods are suboptimal from the information bottleneck perspective and can be sensitive to the pre-trained models.

\vspace{-0.5em}
\section{Conclusion}
\vspace{-0.5em}

We have proposed a new graph classification framework based on graph functional neural process (FNP) and graph generative model. The proposed framework quantifies the predictive uncertainty directly from the functional space and generates the model-level rationales for model interpretability.
The core of our method is that we assume a set of latent rationales which can be mapped into a probabilistic latent space. Motivated by FNP, we design a stochastic correlation matrix to learn the correlations between rationale and graph embeddings. 
The predictive distribution of the graph classifier is conditioned on correlated rationale embeddings and graph embedding. To obtain the graph structures of the rationales, we further incorporate a graph generator into our framework. Extensive experiments on five graph classification datasets demonstrate that our method outperforms state-of-the-art uncertainty quantification  and graph interpretability methods.

\subsubsection*{Acknowledgements}
We would like to thank the anonymous reviewers for their helpful comments. This work was supported in part by NSF (IIS-2008334, IIS-2106961, CAREER IIS-2144338), ONR (MURI N00014-17-1-2656).

\setlength{\itemindent}{-\leftmargin}
\makeatletter
\renewcommand{\@biblabel}[1]{}
\makeatother

\bibliographystyle{plainnat}  
\bibliography{reference}

\appendix
\newpage
\onecolumn

\section{Additional Experiments}\label{apd:exp}

\begin{figure*}[!ht]
	\centering
 	\subfigure[Calibration plot on the Graph-SST2 dataset.]{
		\includegraphics[height=4.5cm]{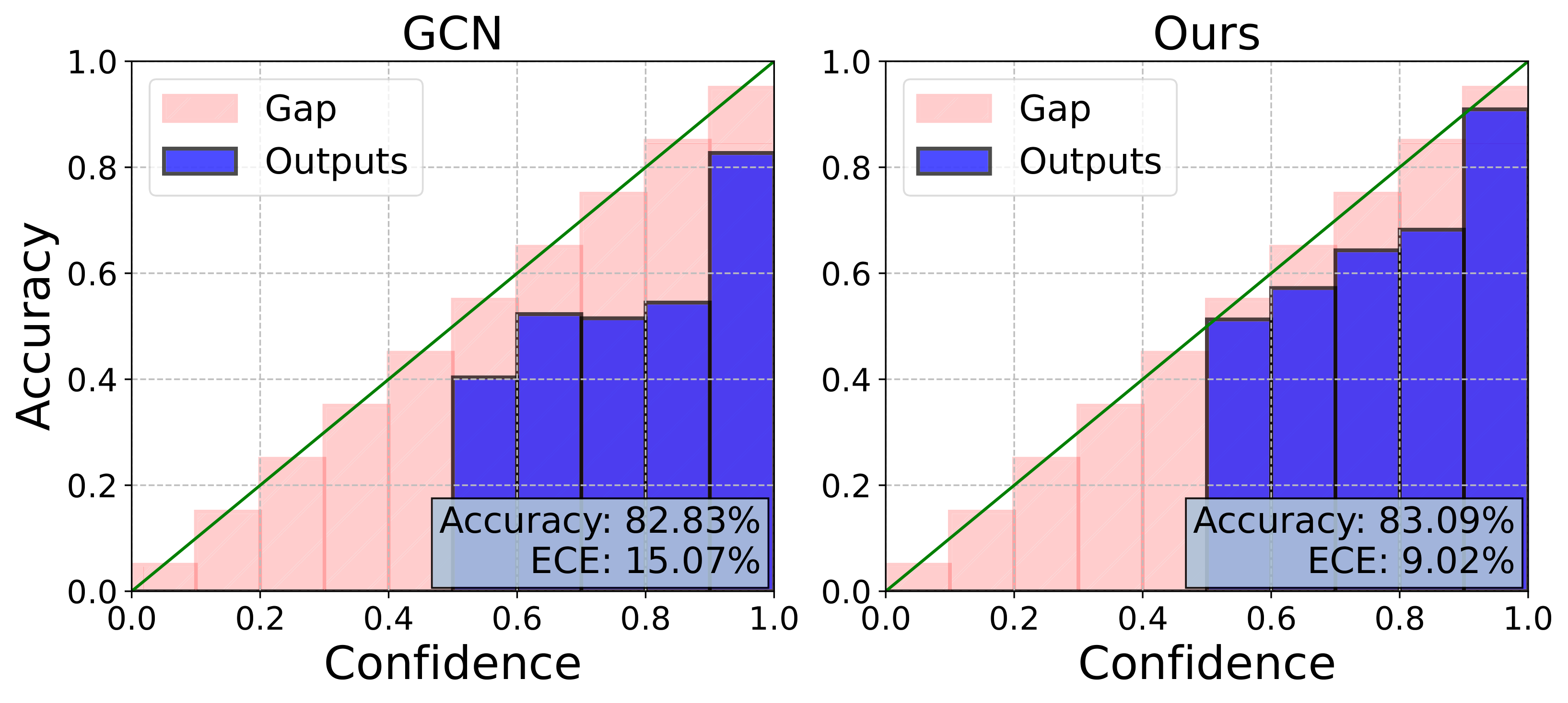}
		\label{fig:calibration}
	}  
	\subfigure[Visualization on the BA2Motifs dataset.]{
		\includegraphics[height=4.5cm]{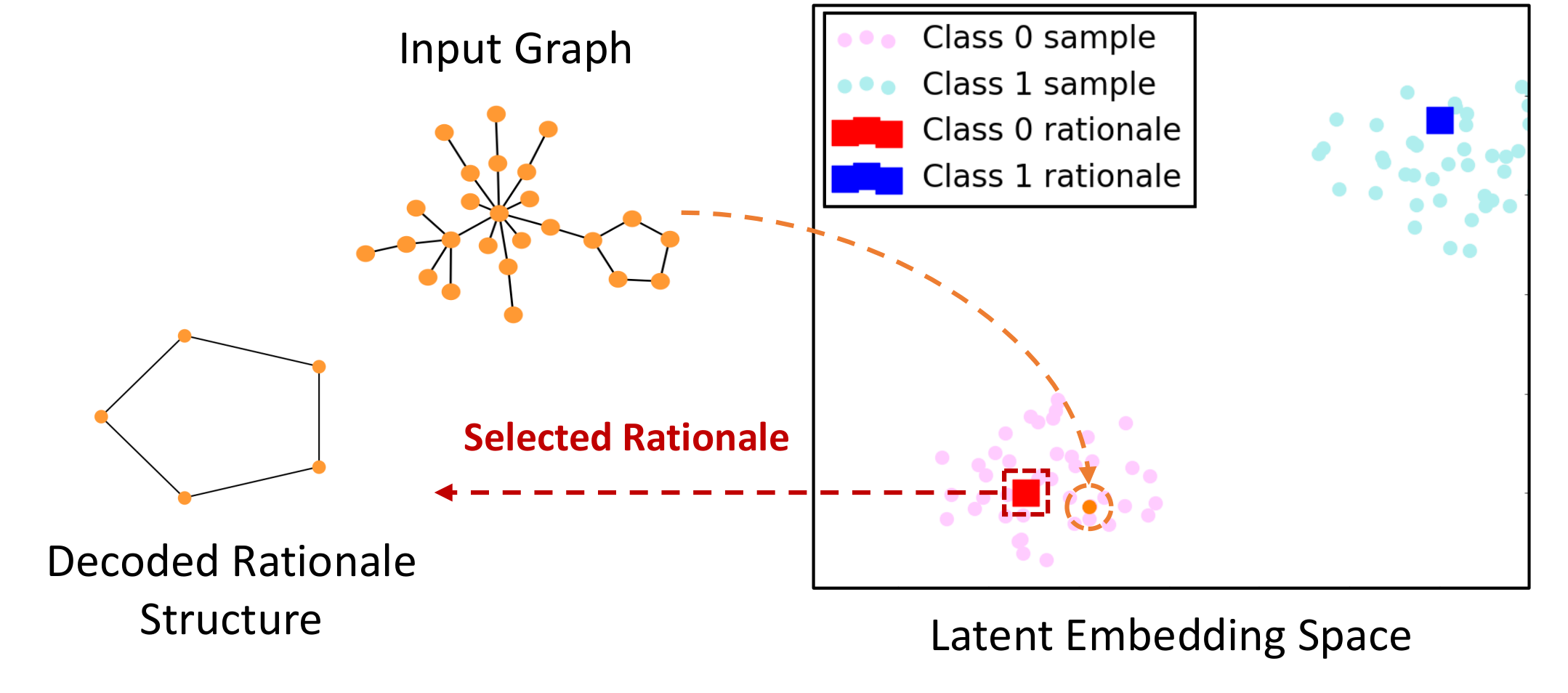}
		\label{fig:test}
	}  
	\caption{(a) \ours improves model calibration significantly while maintaining high predictive performance. (b) \ours learns a set of model-level rationales and decodes the structure of the correlated one for model  interpretability.}\label{fig:intro}
\end{figure*}

\subsection{Fig.~\ref{fig:intro}}

\subsection{Ablation Study}\label{apd:abs}
\begin{table}[h]
  \fontsize{7.5}{9.5}\selectfont \setlength{\tabcolsep}{0.4em}
    \caption{Ablation Study on the BBBP and BACE datasets. We
    use GCN as the backbone. We set $K=1$ when computing RF1.
  }
  \vspace{3ex}
  \centering
  \resizebox{0.46\textwidth}{!}{%
  \small
  \begin{tabular}{@{}ccc|cc|cc@{}}
    \toprule
    & \multicolumn{2}{c|}{ECE} & \multicolumn{2}{c|}{Acc/ROC-AUC} & \multicolumn{2}{c}{RF1} \\ \cmidrule(l){2-7}
    & SST2     & BACE    & SST2         & BACE        & SST2    & BACE    \\ \midrule
    w/o  $\mathbf{z}^D$        &   $9.93$\std{0.92}             & $9.65$\std{2.64}        &  $79.83$\std{1.92}                  & $77.29$\std{3.02}             &  $83.94$\std{0.83}             & $63.12$\std{4.92}       \\
    w/o $\mathbf{C}$ &       $11.78$\std{0.63}         &    $13.82$\std{3.19}     & $82.05$\std{1.37}                   &      $79.23$\std{2.19}       &    $81.25$\std{1.52}           &  $61.58$\std{5.36}       \\
    w/o $\mathbf{z}^R$        &      $15.29$\std{1.53}          &     $16.03$\std{3.74}    &  $82.83$\std{0.53}                  &    $79.78$\std{1.37}         &          NA     &      NA  \\
    w/o EM &$14.65$\std{1.32} & $12.18$\std{2.06}  & $74.12$\std{3.62} & $73.13$\std{1.92} & $75.83$\std{2.39}& $54.13$\std{4.47}\\
   \rowcolor[HTML]{EFEFEF} \ours        & $\bm{9.02}$\std{0.58}               &    $\bm{9.11}$\std{2.96}     &                $\bm{83.08}$\std{0.26}      &     $\bm{80.36}$\std{0.89}      & $\bm{85.85}$\std{0.67}              &   $\bm{66.11}$\std{3.75}      \\ \bottomrule
  \end{tabular}
  }
  \label{table:ablation}
\end{table}

We evaluate the impact of individual components in our method through an ablation study. The results obtained using the GCN architecture on the Graph-SST2 and BACE datasets are summarized in Table~\ref{table:ablation}.

\noindent$\bullet$
The removal of the graph embedding $\mathbf{z}^D$ from the final predictive distribution leads to a significant decline in classification performance. This is due to the fact that without the graph embedding, the model may miss important information unique to the test graph and therefore have a poorer ability to generalize.

\noindent$\bullet$
Removing the binary stochastic correlation matrix $\mathbf{C}$ and using the pairwise kernel distance directly to obtain the local rationale embedding results in a decrease in the model's calibration score. This is because the sampling process of the binary stochastic matrix captures the uncertainty in data correlation, thereby improving the expected calibration error (ECE).

\noindent$\bullet$
The removal of the rationale embedding ($\mathbf{z}^R$) results in a degradation of the Expected Calibration Error (ECE) to the same level as a basic Graph Convolutional Network (GCN) model. This highlights the crucial role the rationale embedding plays in the uncertainty quantification of our method. The RF1 metric cannot be evaluated in this scenario since the rationale embedding is not learned.

\noindent$\bullet$
By eliminating the EM algorithm from the training process, we optimize all model parameters together. We observe a marked decline in performance across all metrics. This is due to the coupling of the rationale and graph embeddings in the same embedding space, making simultaneous optimization difficult.

\subsection{Compatibability with Post-hoc Calibration Method}
\label{post}
In addition, post-hoc calibration techniques aim to enhance model calibration without altering the training procedure of standard neural networks. We evaluate the potential of such methods to improve the performance  of existing uncertainty estimation techniques.Specifically, we apply temperature scaling, a widely used post-hoc calibration, to each model and evaluate Expected Calibration Error (ECE; \citealp{zhuang2023dygen}) on the validation set. Table~\ref{table:ts} shows the results on the Graph-SST2 and BACE datasets. As can be seen, with post-hoc calibration, \ours significantly improves the ECE and remains the best among all methods in terms of uncertainty quantification performance.

\begin{table}[h]
  \caption{ECE with temperature scaling. We report the average performance and standard deviation for 5 random initializations}
  \fontsize{7.5}{9.5}\selectfont \setlength{\tabcolsep}{0.4em}
  \centering
  \vspace{2em}
  \resizebox{0.5\textwidth}{!}{%
  \begin{tabular}{@{}ccc|cc@{}}
    \toprule
    Backbone     & \multicolumn{2}{c|}{GCN} & \multicolumn{2}{c}{GAT} \\ \midrule
    Model        & Graph-SST2     & BBBP    & Graph-SST2    & BBBP    \\ \midrule
    Vanilla      &     $13.34$\std{1.04}        & $15.72$\std{1.8}            &         $11.49$\std{0.51}      & $18.35$ \std{1.84}         \\
    DropEdge     &   $11.95$\std{1.29}           &   $12.25$\std{2.01}        &      $8.02$\std{0.82}         & $16.02$\std{1.96}        \\
    MC-Dropout   &  $11.35$\std{1.24}              &  $13.51$\std{1.73}       &  $10.74$\std{1.21}             & $15.51$\std{3.46}        \\
    SGLD         &      $9.74$\std{0.93}     &     $13.03$\std{2.57}     &      $10.33$\std{1.05}         &       $14.22$\std{2.06}  \\
    BBP          &      $11.95$\std{0.84}          &  $11.12$\std{3.03}       & $8.05$\std{1.57}              &     $14.73$\std{2.27}    \\
    Graph-GP     &   $10.76$\std{1.13}             & $13.12$\std{2.13}        &        $9.27$\std{1.25}       &   $15.01$\std{1.67}      \\
    GSAT         &    $9.69$\std{1.89}            &  $29.95$\std{2.68}    &     $9.28$\std{3.07}            &   $27.70$\std{1.57}     \\
    DeepEnsemble &    $8.26$\std{NA}            & $11.03$\std{NA}         &     $7.43$\std{6.96}          &     $13.74$\std{NA}    \\
    \rowcolor[HTML]{EFEFEF} \ours    &    $7.25$\std{0.74}            &  $\bm{8.72}$\std{1.26}       &        $\bm{6.72}$\std{1.38}       &      $\bm{11.94}$\std{2.97}   \\ \bottomrule
  \end{tabular}
  }
  \label{table:ts}
\end{table}

\subsection{Hyper-parameter Study}
\label{appendix:hyper}
\begin{figure*}
  \centering
  \includegraphics[width=0.95\textwidth]{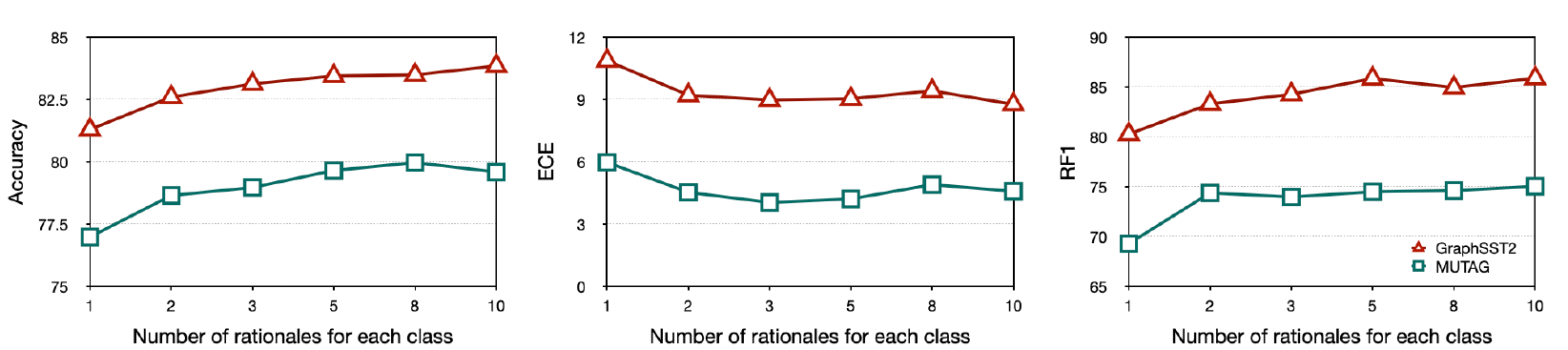}
  \caption{Hyper parameter study on the Graph-SST2 and MUTAG datasets. We use GCN as the backbone and set $K=1$ when computing RF1.}
  \label{fig:hyper}
\end{figure*}

We conduct experiments to study the effect of the number of rationales on the Graph-SST2 and MUTAG datasets. As we can see from Fig. \ref{fig:hyper},
the performance of \ours is stable when $R_k$ is larger than 2. In our experiments, we did not do extensive hyper-parameter tuning and simply set $R_k=5$ for all the datasets.

\section{Additional Related Work}
\noindent\textbf{Uncertainty Quantification}
Bayesian Neural Networks (BNNs) \citep{blundell2015weight, louizos2017multiplicative, mackay1992practical} are realized by first imposing prior distributions over NN parameters, next inferring parameter posteriors and integrating over them to make predictions. However, due to the intractability of posterior inference, approximation methods have been proposed, including variational inference \citep{blundell2015weight, louizos2017multiplicative}, Monte Carlo dropout \citep{gal2016dropout} and stochastic gradient Markov chain Monte Carlo (SG-MCMC) \citep{li2016preconditioned, zhang2019cyclical}. 
Neural Process (NP), a recently introduced framework \citep{garnelo2018neural} attempts to combine the stochastic processes and DNNs. It defines a distribution over a global latent variable to capture the functional uncertainty, while the Functional neural process (FNP) \citep{louizos2019functional} uses a dependency graph to encode the data correlation uncertainty. However, they are both designed for non-graph data. 
Besides the Bayesian methods, model ensembling \citep{lakshminarayanan2017simple} trains multiple DNNs with different initializations and uses their predictions for uncertainty estimation. However, training an ensemble of DNNs can be prohibitively expensive in practice. \citet{kong2020sde} proposes SDE-Net which quantifies uncertainty from a stochastic dynamical system perspective. Recent work by \citet{li2024muben} presents a benchmark for assessing uncertainty quantification in representation learning methods, especially for molecular graphs.

\noindent\textbf{Graph Generation}. 
The initial deep generation models for graphs were autoregressive, such as GraphRNN \citep{you2018graphrnn} and GRAN \citep{liu2019graph} where nodes and edges were generated in a sequential manner. Later, Variational Autoencoder (VAE) based graph generation models were proposed\citep{simonovsky2018graphvae,liu2019graph}. Normalizing flows have also been used for graph generation, with the first application introduced by \citet{liu2019graph}, where a flow models the node representations of a pre-trained autoencoder. Recently, GraphNVP \citep{madhawa2019graphnvp}, and GraphAF \citep{shi2019graphaf} have been proposed for molecular graph generation. GraphNVP consists of two flows, one for the adjacency matrix and another for the node types. GraphAF \citep{shi2019graphaf} is an autoregressive normalizing flow models that samples nodes and edges in sequence. Recently, diffusion models \citep{jo2022score, vignac2023digress, kong2023autoregressive} have been used for graph generation and achieved state-of-the-art performance.

\section{Datasets}
We provide the dataset statistics in Table.~\ref{table:datasets}. All datasets are publicly available:
\begin{enumerate}
    \item BBBP and BACE: \url{https://moleculenet.org/}.

\item Graph-SST2: \url{https://github.com/Graph-COM/GSAT}

\item MUTAG: \url{https://github.com/flyingdoog/PGExplainer}

\item Github: \url{https://chrsmrrs.github.io/datasets/}
\end{enumerate}

\begin{table}[h]
  \caption{Dataset statistics of BBBP, BACE, Graph-SST2, MUTAG, and Github}
  \fontsize{7.5}{9.5}\selectfont \setlength{\tabcolsep}{0.4em}
  \centering
  \vspace{1em}
  \resizebox{0.5\textwidth}{!}{%
  \begin{tabular}{r|ccccc}
    \toprule
     Datasets & BBBP     &  BACE   & Graph-SST2  & MUTAG  & Github   \\ \midrule
    \# of Graphs          & 2039 & 505 & 70042 & 188 & 12725\\
    \# of Nodes (avg) &  24.06 & 36.46 & 10.19  & 17.93 & 52.33 \\
    \# of Edges (avg) &  25.95 & 39.28 & 9.20 & 19.79 & 86.41\\ 
    \# of Nodes (min-max) &2-132 & 10-97&  1-56 & 4-417 & 10-957\\
    \# of Edges (min-max) &1-145 & 10-101&  0-55 & 3-112 & 9-1340\\ \bottomrule
  \end{tabular}
  }
  \label{table:datasets}
\end{table}

\section{Implementation details}
\label{imple}
We adopt ADAM \cite{kingma2014adam} as the optimizer for all the methods and select the learning rate from $\{1\times 10^{-3},5\times 10^{-4}, 1\times 10^{-4} \}$ based on the predictive performance on the validation set. In the generator fine-tuning stage, we adopt a learning rate of $10^{-5}$ for all the datasets.
For a fair comparison, we adopt the same graph convolutional network (GCN)  and graph attention network (GAT) architectures for all the methods as the backbones. We found that the performance of \ours is quite stable with regard to the hyper-parameter $R_k$, \ie the number of rationales for each class, and simply set it as 5 for all the datasets.
For all the datasets, we employ three layers of GCNs with output dimensions equal to 256. We average all node representations as the whole graph representation. The final classifier contains three fully-connected layers in which the hidden dimension is set to 256. The MLPs used in \ours are all one-hidden layers with ReLU activation with hidden dimension 256. 

For all the baselines, we select the hyper-parameters from their recommended ranges in the original papers based on the validation predictive performance. For the GNN interpretability baseline,  GNNExplainer, PGExplainer, and SubgraphX use the implementations provided by Dive into Graphs (DIG) library. GstarX and GSAT use their original code.
\begin{enumerate}
    \item  GNNExplainer, PGExplainer, and SubgraphX (based on DIG): \url{https://github.com/divelab/DIG}.

\item GstarX: \url{https://github.com/ShichangZh/GStarX}

\item GSAT: \url{https://github.com/Graph-COM/GSAT}

\end{enumerate}
For XGNN, their original code cannot be used in the Pytorch-geometric framework. We reimplement it using the PyTorch-Geometric framework under the guidance of their paper and code in DIG library \url{https://github.com/divelab/DIG/blob/main/dig/xgraph/XGNN/gcn.py}.

All experiments are conducted on CPU: Intel(R) Core(TM) i7-5930K CPU @ 3.50GHz and GPU: NVIDIA GeForce RTX A5000 GPUs using python 3.8 and PyTorch 1.12.

\end{document}